\documentclass[hf]{ceurart}  %

\usepackage[frozencache,cachedir=minted-cache]{minted}
\setminted{
    linenos,
    breaklines=true,
    fontfamily=courier
}
\usepackage{xpatch}
\makeatletter
\def\disablefcolorbox{\renewcommand\fcolorbox[4][]{##4}}
\xpatchcmd{\inputminted}{\minted@fvset}{\minted@fvset\disablefcolorbox}{}{}
\makeatother

\usepackage{comment}
\usepackage{placeins}
\usepackage[compatibility=false]{caption}

\usepackage{cleveref}

\newcommand{\furl}[1]{\footnote{\scriptsize \url{#1}}}

\usepackage{svg}

\usepackage[breakable,skins]{tcolorbox}
\newcommand{\llmprompt}[1]{\begin{tcolorbox}[width=\columnwidth,colframe=gray,arc=2mm,breakable,pad at break*=1mm,enhanced jigsaw] \footnotesize\textbf{User:} #1 \end{tcolorbox}}
\newcommand{\llmanswer}[1]{\begin{tcolorbox}[width=\columnwidth,colframe=gray,colback=green!40,arc=2mm,breakable,pad at break*=1mm,enhanced jigsaw] \footnotesize\textbf{GPT-4:} #1 \end{tcolorbox}}

\begin{document}

\copyrightyear{2024}
\copyrightclause{Copyright for this paper by its authors.}%

\conference{Preprint. Under review.}

\title{Towards Enabling FAIR Dataspaces Using Large Language Models}
\author[1,2]{Benedikt T. Arnold}[%
orcid=0000-0001-8594-880X,
email=benedikt.arnold@fit.fraunhofer.de
]
\author[1,2]{Johannes Theissen-Lipp}[orcid=0000-0002-2639-1949]
\author[1,3]{Diego Collarana}[orcid=0000-0002-2583-0778]
\author[1,2]{Christoph Lange}[orcid=0000-0001-9879-3827]
\author[1,2]{Sandra Geisler}[orcid=0000-0002-8970-6282]
\author[4]{Edward Curry}[orcid=0000-0001-8236-6433]
\author[1,2]{Stefan Decker}[orcid=0000-0001-6324-7164]

\address[1]{Fraunhofer Institute for Applied Information Technology FIT, St. Augustin, Germany}
\address[2]{RWTH Aachen University, Aachen, Germany}
\address[3]{Universidad Privada Boliviana, Bolivia}
\address[4]{University of Galway, Galway, Ireland}
\begin{abstract}
Dataspaces have recently gained adoption across various sectors, including traditionally less digitized domains such as culture. Leveraging Semantic Web technologies helps to make dataspaces FAIR, but their complexity poses a significant challenge to the adoption of dataspaces and increases their cost. The advent of Large Language Models (LLMs) raises the question of how these models can support the adoption of FAIR dataspaces. In this work, we demonstrate the potential of LLMs in dataspaces with a concrete example. We also derive a research agenda for exploring this emerging field.
\end{abstract}

\begin{keywords}
    Dataspaces \sep
    FAIR Data Principles \sep
    Large Language Models 
\end{keywords}

\maketitle

\section{Introduction}\label{sec:intro}
In recent years, many dataspace verticalizations have emerged from European dataspace initiatives such as the \emph{International Data Spaces (IDS)}~\cite{bader2020international} and \emph{Gaia-X}\footnote{\url{https://gaia-x.eu/}} in different domains. Dataspaces are now also being established in domains with less affinity to digitization, such as culture. The strong grounding in Semantic Web technology (in the following also: semantics) promotes the realization of the \emph{FAIR} principles in dataspaces. However, this raises the need for adopters of dataspaces to familiarize themselves with these technologies.
This is a significant entry barrier and incurs costs for learning semantics or outsourcing the related tasks. The recent advent and broad adoption of \emph{Large Language Models (LLMs)} raises the question of how this technology can be sensibly used to ease the adoption of FAIR dataspaces at a lower cost. %
In this work, we demonstrate the potential of LLMs for dataspaces with an example in~\cref{sec:gpt4atwork}. In~\cref{sec:agenda}, we propose a research agenda for investigating LLMs in FAIR dataspaces. Before, we introduce some important concepts:

\paragraph{Dataspaces}%
The term \emph{dataspace} was coined in 2005 %
and has evolved since then, with scientific definitions gathered in~\cite{theissen2023semantics}. In this work, we refer to dataspaces as \emph{a multi-sided data platform} connecting participants in an ecosystem~\citep{otto2019designing}. Importantly, dataspaces do not incorporate an integration layer to bridge heterogeneity; instead, each data source remains unaltered, eliminating the need to transfer data to a centralized storage location.
This omission of the integration %
aspect in data exchange reduces the initial workload and concentrates on an \emph{as needed} best-effort strategy for data integration~\cite{curry2020future}. Significantly, the decentralized storage concept benefits dataspace participants as they maintain sovereignty over their data~\cite{geisler2021knowledge}.
Dataspaces commonly use semantics to ensure a ``common language'' and foster FAIR data sharing~\citep{theissen2023semantics}.

\paragraph{FAIR Principles}
The FAIR Data Principles\footnote{\url{https://www.go-fair.org/fair-principles/}} %
aim at supporting the \emph{Findability}, \emph{Accessibility}, \emph{Interoperability}, and \emph{Reusability} of data
and have gained widespread adoption. 
FAIR comprises these technical prerequisites: persistent identifiers (PIDs), rich metadata and open protocols -- of which PIDs and rich metadata are closely linked with semantics~\citep{theissen2023semantics}. In this paper, we highlight some of the most complicated and tedious associated tasks, which are: Enhancing existing semantic metadata schemas, creating instances from these schemas and understanding semantic data. %
We see a high potential to tackle these tasks more efficiently using LLMs -- fostering the incorporation of the FAIR principles if done correctly. %

\paragraph{Large Language Models}%
Generative AI, in the form of Large Language Models (LLMs), represents a fundamental shift in text processing and knowledge generation methodologies. LLMs have shown excellent capabilities in understanding and generating human-like text, leading to breakthroughs in various downstream applications, including text summarization, content generation, and conversational systems~\cite{lewis2020retrieval}.
LLMs such as GPT-4~\citep{openai2023gpt4} or Mixtral 8x7B\footnote{\url{https://mistral.ai/news/mixtral-of-experts/}} predict the next tokens in an auto-regressive manner, given an input sequence (prompt). They are first trained on a massive and diverse dataset in an unsupervised manner to produce base models and subsequently fine-tuned on a smaller and more specific dataset, aiming at performance improvement in specific tasks, like instruction-following~\cite{lewis2020retrieval}.

\section{Tasks in FAIR Dataspaces and the Potential of LLMs}\label{sec:gpt4atwork}
With a practical example, we demonstrate how tasks related to providing and consuming FAIR data via dataspaces can be aided by LLMs, in here GPT-4. Following this proof-of-concept, we outline paths to solve these tasks more sophisticatedly for more realistic cases in \cref{sec:agenda}.

Rich metadata benefits especially \emph{Findability} and \emph{Interoperability}. This can be achieved using semantics (cf.~\cref{sec:intro}). For simplicity, consider the specification of an offered dataset in \cref{listing:shacl_base}. In the listing, two \emph{SHACL (Shapes Constraint Language)}\footnote{\url{https://www.w3.org/TR/shacl/}} shapes ensure that each dataset has a title and a usage policy. The policy is needed to make the dataset a valid dataspace offering for other participants to conclude a usage contract.
Our example considers the cultural domain and how digitized paintings can be offered. A relevant ontology in the cultural domain is the \emph{Integrated Authority File (GND)}, associated with a \emph{Knowledge Graph (KG)} that represents ``the largest collection of cultural and research authority data in the German-speaking countries''\footnote{\url{https://gnd.network/Webs/gnd/EN/Home/home_node.html}}. %
We start our example with a museum curator who wants to offer her data in the dataspace and is the first to do so from the art domain. Therefore, they must extend the existing metadata schema with corresponding properties. The curator is familiar with domain-specific metadata standards but wants to focus on the concepts instead of syntactical subtleties of formats such as \emph{Turtle}\footnote{\url{https://www.w3.org/TR/turtle/}}. She knows that a painter can be represented as a painting's \mintinline{sparql}{gndo:firstArtist} that is a \mintinline{sparql}{gndo:DifferentiatedPerson}, which has the property \mintinline{sparql}{gndo:gndIdentifier} that provides the person's GND id. \texttt{gndo} denotes the GND namespace. Using this id is important as it can serve as a unique \emph{Persistent Identifier} (PID) for this person and thus supports the %
FAIR principles.
The property \mintinline{sparql}{gndo:dateOfProduction} gives the painting's creation date.
With this \emph{domain knowledge}, we employ the following prompt: %
\llmprompt{
    As a Semantic Web expert, please extend the SHACL shapes given below such that I can also describe digital versions of paintings with them, using properties from the Integrated Authority File, the GND. For the painter, please use a `gndo:firstArtist' that is a `gndo:DifferentiatedPerson' which has its GND id as a property. For the year of creation, please use `gndo:dateOfProduction'.

    \texttt{\`{}\`{}\`{}turtle ...}
}

\begin{figure}[p]  %
\noindent\begin{minipage}[t]{.45\textwidth}
\inputminted[fontsize=\tiny,escapeinside=||]{sparql}{./listings_example/base_reduced.ttl}
\captionof{listing}{The base SHACL shapes.}
\label{listing:shacl_base}
\vskip12pt
\inputminted[fontsize=\tiny,escapeinside=||]{sparql}{./listings_example/extended_cleaned_concise_reduced.ttl}
\addtocounter{listing}{1}
\captionof{listing}{The corrected shapes from \cref{listing:shacl_extended}.}
\label{listing:shacl_extended_cleaned}
\addtocounter{listing}{-2}
\end{minipage}\hfill
\begin{minipage}[t]{.45\textwidth}
\inputminted[fontsize=\tiny,escapeinside=||]{sparql}{./listings_example/extended_concise_reduced.ttl}
\captionof{listing}{The SHACL shapes with extensions\\related to paintings, by GPT-4.}
\label{listing:shacl_extended}
\addtocounter{listing}{1}
\end{minipage}
\end{figure}

Note that we have not included the exact identifier for the GND id in the prompt.
The response of GPT-4, with surrounding explanatory comments omitted, are the shapes presented in~\cref{listing:shacl_extended} -- importantly, the existing shapes were not harmed. The painter with their identifier and the date of production have been correctly added.
However, the painter is also referenced with their \mintinline[escapeinside=||]{sparql}{gndo:preferredName|\allowbreak|ForThePerson} (lines 17--22), which was not intended. Besides, the \mintinline{sparql}{gndo:dateOfProduction}'s datatype (line 33) is either an \mintinline{sparql}{rdfs:Literal} or \mintinline{sparql}{xsd:dateTime}. Hence, we ask GPT-4 to correct these mistakes using a subsequent prompt:
\llmprompt{
    Please do not include the preferred name of the painter, and consider that the dateOfProduction should be of type `xsd:dateTime'.
}

The resulting shapes~(\cref{listing:shacl_extended_cleaned}) again remain consistent with their previous versions, and the issues are correctly resolved.
To support people new to the metadata schema and semantics, GPT-4 can also be used to visualize it, e.g., in \emph{UML}, using a text-based notation such as \emph{PlantUML}\footnote{\url{https://plantuml.com/}}.%

Besides creating the initial schema, an important task is creating the instances based on the schema. We use the real example of Caspar David Friedrich's painting ``Der Wanderer über dem Nebelmeer'', which was painted around 1818, according to the GND\footnote{\url{https://d-nb.info/gnd/1069695750/about/lds.ttl}}.
For this, we can make use of the OpenAI API's \emph{Function Calling} feature\footnote{\url{https://platform.openai.com/docs/guides/function-calling?lang=python}}. This allows GPT-4 to invoke externally defined functions in a two-turn interaction. We define a function to extract the id for a painter from the GND given the name with a \emph{SPARQL} query. Based on the following prompt, we let GPT-4 create the instance, using the function if needed:
\llmprompt{
    You are given the following schema:\newline
\`{}\`{}\`{}turtle\newline
... omitted for conciseness ...\newline
\`{}\`{}\`{}\newline
Please create an instance of it for the painting ``Der Wanderer über dem Nebelmeer'' by Caspar David Friedrich, which was created in 1818. You may look up the GND ID of the artist if you need to.
}

First, the supplied function is invoked with the painter's name. Subsequently, the instance in~\cref{listing:instance} is generated, alongside a description of how GPT-4 resolved prompt ambiguities:%
\llmanswer{
Please note that:\newline
- I've used \`{}\@de\`{} language tag with the title to specify the language is German.\newline
- For the \`{}odrl:hasPolicy\`{}, I created a placeholder URI (\`{}http://example.org/policy/12345\`{}) as the policy detail wasn't specified.\newline
- The date of production \`{}1818-01-01\`{} is set to January 1st, 1818, as the exact date wasn't provided. In this example, the time component is set to midnight. Adjust the date according to specific requirements or available data.
}

We generate the usage policy based on the \emph{Open Digital Rights Language (ODRL)}\footnote{\url{https://www.w3.org/TR/odrl-model/}}, following the approach presented by~\citet{mustafa2024from}, who supplement GPT-4 with a sophisticated description of the key parts of the ODRL ontology%
. We provide the following prompt:
\llmprompt{
Given the following instance, create an ODRL policy that allows the dataset to be used within Germany until 2024-05-10. ... omitted for conciseness ...
}
\Cref{listing:policy} shows the resulting usage policy, preserving the policy's \emph{Internationalized Resource Identifier (IRI)} from~\cref{listing:instance}.

\begin{figure}[p]
\noindent\begin{minipage}[b]{.45\textwidth}
\inputminted[fontsize=\tiny]{sparql}{./listings_example/instance.ttl}
\captionof{listing}{The instance as created by GPT-4 based\\on the description prompt and the GND lookup.}
\label{listing:instance}
\end{minipage}\hfill
\begin{minipage}[b]{.45\textwidth}
\inputminted[fontsize=\tiny]{sparql}{./listings_example/policy.ttl}
\captionof{listing}{The policy generated for the instance\\in \cref{listing:instance}. The IRI has been preserved.}
\label{listing:policy}
\end{minipage}
\end{figure}

\newpage
Finally, we start a new conversation with GPT-4 and ask it to, given the SHACL shapes, explain the created instance, together with its associated usage policy and get this result:
\llmanswer{
This set of information is essentially a structured way to describe a dataset, in this case, related to a digital representation of a painting. Here's the breakdown:\newline
1. Dataset Identification: The data snippet describes a dataset named ``Der Wanderer über dem Nebelmeer'' which is German for ``The Wanderer above the Sea of Fog.'' This dataset is identified on the web at ex:DerWandererÜberDemNebelmeer.\newline
2. Title: The title of the dataset is given as ``Der Wanderer über dem Nebelmeer,'' which aligns with the actual painting's title.\newline
3. Policy Association: The dataset is associated with a policy indicated by the URI <http://example.org/\allowbreak policy/12345>. This policy defines how the dataset can be used or accessed.\newline
4. Artist Information:
\begin{itemize}
    \item The dataset specifies an artist associated with it, identified under the system gndo:firstArtist.
    \item This artist is a "Differentiated Person" meaning they are uniquely identified, not just a name but a specific individual.
    \item The artist has a unique identifier "118535889," which helps in uniquely identifying and retrieving information about the artist from a database or registry.
\end{itemize}
5. Date of Production: The dataset includes a production date, 1818-01-01T00:00:00 formatted in the standard date-time format. This likely refers to the creation date of the painting.\newline
6. Policy Details:
\begin{itemize}
    \item The dataset is associated with a specific usage policy (ex:policy12345), which is an object defining permissions related to the dataset.
    \item This policy allows the action of use on the dataset but has constraints:
    \begin{itemize}
        \item It can only be used within Germany (``DE''), indicated by the spatial constraint.
        \item It must be used before 2024-05-10T23:59:59, as indicated by the dateTime constraint.
    \end{itemize}
\end{itemize}
This structured format helps machines and applications understand and process detailed metadata about digital resources, ensuring they're used and shared according to specified rules and providing essential context such as creation details and usage rights.
}

This is mostly a valid description of the instance. However, some important details get lost, including the reference to the source of the object's id, the GND.

In summary, we have shown that GPT-4 can help with some of the key tasks for preparing data for a FAIR exchange using dataspaces. The example was intentionally simple, and the LLM outputs had some issues. Still, it shows the general applicability of LLMs in this domain. In the next section, we propose a research agenda for using open LLMs in real dataspace settings, in combination with KGs to address some of the LLMs' intrinsic limitations.

\section{Research Agenda}\label{sec:agenda}
As \citet{fernandez2023large} point out, LLMs have drastically changed the capabilities of \emph{Natural Language Processing} systems as they can generate text and code of viable quality based on a natural language interface--with a better understanding of the underlying data's semantics. %
We are convinced that LLMs can aid many tasks in dataspaces, enhancing their adoption and FAIR\-ness. %
In this section, we provide a research agenda, with research questions across multiple areas related to dataspaces that we consider pivotal %
for assessing how LLMs can benefit dataspaces. %

\paragraph{Interactive and Automated Systems}
LLMs can be used in conversational settings and as prompt-based generators, raising the question of how assisting systems should be designed: interactive or as automated background services. We believe that this depends on the context.

One example of an automated system could be a dataspace metadata broker like the \emph{Federated Catalogue} in Gaia-X. The stored \emph{Self-Descriptions} in RDF can automatically be enhanced with an explanation by a specifically prompted LLM that can access the ontologies underlying the catalogue, see \emph{Integration of Knowledge and Correctness}.
Besides, mapping user data schemas to common standards can be automated using LLMs, fostering FAIRness~\citep{sundaram2023making}. %

In recent unpublished work, we use an interactive method to create Self-Descriptions with a natural language interface, leveraging conventional technology: Users provide attributes conversationally, and can ask questions. The system auto-corrects inputs using known facts and generates the self-description with standard RDF software, thus avoiding syntax errors.

For interactive approaches, it will be crucial to determine how the system has to be designed to be user-friendly. %
For automated approaches, ways have to be found to deal with unexpected output from the LLM, i.e., both hallucinations and unmatched output format requirements. For the latter, token sampling based on formal grammars~\citep{geng2023grammar} is an interesting research direction. %

\paragraph{Adaptation: Prompt Engineering and Fine-Tuning}
When it comes to adapting a model to a new context like dataspaces and new tasks, two options are possible: The first technique is called \emph{Prompt Engineering} %
and describes crafting specific prompts to obtain the desired output. Typically, there is additional context or example outputs given in the prompt that the LLM can use in addition to the knowledge encoded in its parameters.
Another approach for adaption is \emph{fine-tuning} where knowledge and desired output format are directly encoded into the parameters of the LLM, efficiently possible with adapter parameters~\cite{dettmers2024qlora}. For this, a dataset has to be carefully crafted, and enough data has to be available.

Both approaches differ cost- and time-wise. Fine-tuning incurs an initial additional cost for constructing the dataset and performing the resource-intense fine-tuning process. %
Prompt Engineering does not add cost in the beginning apart from building the prompt but requires more time during inference to process the sophisticated, context-enhanced prompts.
Both approaches are worth investigating in the context of dataspaces: Few-shot prompting can be used to obtain the desired results when data scarcity or restrictive usage policies prohibit fine-tuning. However, fine-tuning is of particular interest in generating structured data, e.g., in JSON or Turtle, and can be required to make smaller LLMs adhere to the syntax.

\paragraph{Integration of Knowledge and Correctness}
Due to the intrinsic limitations of LLMs, i.e., hallucination, expensive updates, and lack of provenance, extensive research is carried out on interfacing them with KGs~\cite{collarana2024knowledge}.
This is possible based on a more sophisticated variant of Prompt Engineering, \emph{Retrieval Augmented Generation (RAG)}~\citep{lewis2020retrieval}. In RAG, the prompt is encoded using an embedding model and matched against the embeddings of knowledge in the form of documents saved in vector stores. The most similar results are then injected into the prompt. KGs can also be encoded into embeddings, making the knowledge stored in them accessible for LLMs. For outputs generated from injected KG data, the source part of the graph can be given as provenance. This way, users can even verify the results.

More broadly, \citet{pan2024unifying} present a roadmap summarizing the possibilities for interfacing both technologies. %
The integration of KGs and LLMs is pivotal, especially in the context of FAIR dataspaces, as the metadata exchange in dataspaces is KG-based, especially in IDS and Gaia-X. Exploiting them as a source of knowledge can make LLMs more reliable for assisting dataspace tasks.
Besides, open alternatives to OpenAI's function calling, like agent settings, are worth investigating for interfacing with data sources, e.g. for PID retrieval.

\paragraph{Open Models for Data Sovereignty}
In our example in \cref{sec:gpt4atwork}, we used the highly advanced but proprietary LLM GPT-4 by OpenAI to demonstrate general applicability. While valid for examples, two main reasons make using GPT-4 in dataspaces problematic: First, its proprietary nature creates a reliance on the user for the availability of OpenAI's API. Software built upon this API becomes immediately useless once availability is not given. As the effectiveness of prompts is often highly specific to the LLM at hand, there is a significant vendor lock-in. %
Besides, data sovereignty is one of the core selling points of dataspaces, i.e., the data stays in the participants' infrastructures and is exchanged peer-to-peer. This contradicts the widespread adoption of closed models like GPT-4 for dataspaces, which require sending sensitive data to a third-party service. For GDPR-relevant data, this can even be illegal.
For these reasons, freely available LLMs should be preferred, although their performance does not yet match that of GPT-4. Multiple models with different strengths and vastly differing sizes have emerged from the small \emph{Phi-2}\footnote{\url{https://www.microsoft.com/en-us/research/blog/phi-2-the-surprising-power-of-small-language-models/}} (2.7B param.) to large models like the aforementioned \emph{Mixtral-8x7B} (46.7B param.).
For this to be feasible, we propose to investigate the following research questions:
\begin{itemize}
    \item Given an application domain and a specific LLM use case: What is a reasonable tradeoff between model size (i.e., inference cost), fine-tuning effort, model performance and safety?
    \item How can synergies between dataspace participants be used to fine-tune models? %
    \item Where is using GPT-4 appropriate and its benefits outweigh the downsides?
    \item How can dataspace participants be empowered to perform inference on the edge with equal or less (personnel) cost compared to the hosted OpenAI API?
\end{itemize}

\paragraph{Efficiency And Latency}  %
Energy efficiency and latency are major aspects for self-hosted LLMs and out of control for closed models like GPT-4. Since the widespread interest in LLMs has grown, performance has drastically improved, e.g. with \emph{QLoRA}~\cite{dettmers2024qlora} for fine-tuning. %
Quantization, i.e. representing the LLM parameters with fewer than the canonical 16 bits, strongly reduces the required amount of RAM and the inference latency, allowing for larger models on less powerful hardware, even CPUs, at often marginally reduced performance~\citep{shen2023efficient}. %

Subsequent research in the context of dataspaces will focus on how those scientific and engineering advances can be translated into dataspace-related tools. Specifically, we believe it is crucial to determine how the dataspace participants' acceptance of LLM-based tools is affected by latency, energy consumption, and the balance of latency and output quality. 

\paragraph{Safety}
LLMs generate text based on their internal parameters, optimized on large amounts of text, cf.\ \cref{sec:intro}. Therefore, the outputs reflect what the model has ``seen'' during training, leading to the adoption of biases in the used data--which is problematic for marginalized groups in particular, in-depth analyzed by~\citet{bender2021dangers}. This is especially critical because, for most LLMs, the underlying training data is not openly available for review.

It is pivotal for our research to keep safety in mind and be careful while developing and deploying LLMs for dataspace applications. As an overarching consideration for our research, awareness of this topic and the potential problems must be addressed. Unified data-model life-cycles are needed to ensure the provenance and trust in the data used to train LLMs~\cite{curry2023foundation}. Guardrails must be implemented to avoid problematic LLM generations being displayed to users or used in automatic systems. However, this must be balanced with preserving the system's core functionality. The integration of factual knowledge with provenance using KGs aims at this. %
To ensure safety, the actual dataspace participants must be involved during research, e.g., through user studies, to detect problematic cases.

\begin{acknowledgments}
    This work has been partially funded by the German Federal Ministry of Education and Research (BMBF) under grant number FAIRDS05, by the German Federal Ministry for Economic Affairs and Climate Action (BMWK) under grant number 01MJ22011B, by the German Federal Government Commissioner for Culture and the Media (BKM) under grant number 2522DIG012 and with financial support of Science Foundation Ireland (SFI) under grant number SFI/12/RC/2289\_P2.
\end{acknowledgments}

\FloatBarrier
\newpage

\bibliography{bibliography}

\begin{thebibliography}{17}
\expandafter\ifx\csname natexlab\endcsname\relax\def\natexlab#1{#1}\fi
\providecommand{\url}[1]{\texttt{#1}}
\providecommand{\href}[2]{#2}
\providecommand{\path}[1]{#1}
\providecommand{\DOIprefix}{doi:}
\providecommand{\ArXivprefix}{arXiv:}
\providecommand{\URLprefix}{URL: }
\providecommand{\Pubmedprefix}{pmid:}
\providecommand{\doi}[1]{\href{http://dx.doi.org/#1}{\path{#1}}}
\providecommand{\Pubmed}[1]{\href{pmid:#1}{\path{#1}}}
\providecommand{\bibinfo}[2]{#2}
\ifx\xfnm\relax \def\xfnm[#1]{\unskip,\space#1}\fi
\bibitem[{Bader et~al.(2020)Bader, Pullmann, Mader, Tramp, Quix, M{\"u}ller,
  Aky{\"u}rek, B{\"o}ckmann, Imbusch, Lipp, Geisler, and
  Lange}]{bader2020international}
\bibinfo{author}{S.~Bader}, \bibinfo{author}{J.~Pullmann},
  \bibinfo{author}{C.~Mader}, \bibinfo{author}{S.~Tramp},
  \bibinfo{author}{C.~Quix}, \bibinfo{author}{A.~W. M{\"u}ller},
  \bibinfo{author}{H.~Aky{\"u}rek}, \bibinfo{author}{M.~B{\"o}ckmann},
  \bibinfo{author}{B.~T. Imbusch}, \bibinfo{author}{J.~Lipp},
  \bibinfo{author}{S.~Geisler}, \bibinfo{author}{C.~Lange},
\newblock \bibinfo{title}{{The International Data Spaces Information Model --
  An Ontology for Sovereign Exchange of Digital Content}},
\newblock in: \bibinfo{booktitle}{ISWC}, \bibinfo{year}{2020}.
\bibitem[{Theissen-Lipp et~al.(2023)Theissen-Lipp, Kocher, Lange, Decker,
  Paulus, Pomp, and Curry}]{theissen2023semantics}
\bibinfo{author}{J.~Theissen-Lipp}, \bibinfo{author}{M.~Kocher},
  \bibinfo{author}{C.~Lange}, \bibinfo{author}{S.~Decker},
  \bibinfo{author}{A.~Paulus}, \bibinfo{author}{A.~Pomp},
  \bibinfo{author}{E.~Curry},
\newblock \bibinfo{title}{{Semantics in Dataspaces: Origin and Future
  Directions}},
\newblock {WWW} '23 Companion, \bibinfo{year}{2023}.
\bibitem[{Otto and Jarke(2019)}]{otto2019designing}
\bibinfo{author}{B.~Otto}, \bibinfo{author}{M.~Jarke},
\newblock \bibinfo{title}{{Designing a multi-sided data platform: findings from
  the International Data Spaces case}},
\newblock \bibinfo{journal}{Electronic Markets} \bibinfo{volume}{29}
  (\bibinfo{year}{2019}).
\bibitem[{Curry(2020)}]{curry2020future}
\bibinfo{author}{E.~Curry}, \bibinfo{title}{Future Research Directions for
  Dataspaces, Data Ecosystems, and Intelligent Systems},
  \bibinfo{publisher}{Springer International Publishing},
  \bibinfo{address}{Cham}, \bibinfo{year}{2020}.
\bibitem[{Geisler et~al.(2021)Geisler, Vidal, Cappiello, L\'{o}scio, Gal,
  Jarke, Lenzerini, Missier, Otto, Paja, Pernici, and
  Rehof}]{geisler2021knowledge}
\bibinfo{author}{S.~Geisler}, \bibinfo{author}{M.-E. Vidal},
  \bibinfo{author}{C.~Cappiello}, \bibinfo{author}{B.~F. L\'{o}scio},
  \bibinfo{author}{A.~Gal}, \bibinfo{author}{M.~Jarke},
  \bibinfo{author}{M.~Lenzerini}, \bibinfo{author}{P.~Missier},
  \bibinfo{author}{B.~Otto}, \bibinfo{author}{E.~Paja},
  \bibinfo{author}{B.~Pernici}, \bibinfo{author}{J.~Rehof},
\newblock \bibinfo{title}{Knowledge-driven data ecosystems toward data
  transparency},
\newblock \bibinfo{journal}{Journal of Data and Information Quality}
  \bibinfo{volume}{14} (\bibinfo{year}{2021}).
\bibitem[{Lewis et~al.(2020)Lewis, Perez, Piktus, Petroni, Karpukhin, Goyal,
  K{\"u}ttler, Lewis, Yih, Rockt{\"a}schel et~al.}]{lewis2020retrieval}
\bibinfo{author}{P.~Lewis}, \bibinfo{author}{E.~Perez},
  \bibinfo{author}{A.~Piktus}, \bibinfo{author}{F.~Petroni},
  \bibinfo{author}{V.~Karpukhin}, \bibinfo{author}{N.~Goyal},
  \bibinfo{author}{H.~K{\"u}ttler}, \bibinfo{author}{M.~Lewis},
  \bibinfo{author}{W.-t. Yih}, \bibinfo{author}{T.~Rockt{\"a}schel}, et~al.,
\newblock \bibinfo{title}{{Retrieval-Augmented Generation for
  Knowledge-Intensive NLP Tasks}},
\newblock \bibinfo{journal}{NeurIPS}  (\bibinfo{year}{2020}).
\bibitem[{OpenAI(2023)}]{openai2023gpt4}
\bibinfo{author}{OpenAI}, \bibinfo{title}{{GPT-4} technical report},
  \bibinfo{year}{2023}. \href{http://arxiv.org/abs/2303.08774}{{\tt
  arXiv:2303.08774}}.
\bibitem[{Mustafa et~al.(2024)Mustafa, Arnold, Collarana, Nadgeri, Quix, Lange,
  and Decker}]{mustafa2024from}
\bibinfo{author}{D.~M. Mustafa}, \bibinfo{author}{B.~T. Arnold},
  \bibinfo{author}{D.~Collarana}, \bibinfo{author}{A.~Nadgeri},
  \bibinfo{author}{C.~Quix}, \bibinfo{author}{C.~Lange},
  \bibinfo{author}{S.~Decker},
\newblock \bibinfo{title}{{From Instructions to ODRL Usage Policies: An
  Ontology Guided Approach}},
\newblock \bibinfo{journal}{under review}  (\bibinfo{year}{2024}).
\bibitem[{Fernandez et~al.(2023)Fernandez, Elmore, Franklin, Krishnan, and
  Tan}]{fernandez2023large}
\bibinfo{author}{R.~C. Fernandez}, \bibinfo{author}{A.~J. Elmore},
  \bibinfo{author}{M.~J. Franklin}, \bibinfo{author}{S.~Krishnan},
  \bibinfo{author}{C.~Tan},
\newblock \bibinfo{title}{How large language models will disrupt data
  management},
\newblock \bibinfo{journal}{Proceedings of the VLDB Endowment}
  \bibinfo{volume}{16} (\bibinfo{year}{2023}).
\bibitem[{Sundaram and Musen(2023)}]{sundaram2023making}
\bibinfo{author}{S.~S. Sundaram}, \bibinfo{author}{M.~A. Musen},
  \bibinfo{title}{{Making Metadata More FAIR Using Large Language Models}},
  \bibinfo{year}{2023}. \href{http://arxiv.org/abs/2307.13085}{{\tt
  arXiv:2307.13085}}.
\bibitem[{Geng et~al.(2023)Geng, Josifoski, Peyrard, and
  West}]{geng2023grammar}
\bibinfo{author}{S.~Geng}, \bibinfo{author}{M.~Josifoski},
  \bibinfo{author}{M.~Peyrard}, \bibinfo{author}{R.~West},
\newblock \bibinfo{title}{{Grammar-constrained decoding for structured NLP
  tasks without finetuning}},
\newblock in: \bibinfo{booktitle}{EMNLP}, \bibinfo{publisher}{ACL},
  \bibinfo{year}{2023}.
\bibitem[{Dettmers et~al.(2024)Dettmers, Pagnoni, Holtzman, and
  Zettlemoyer}]{dettmers2024qlora}
\bibinfo{author}{T.~Dettmers}, \bibinfo{author}{A.~Pagnoni},
  \bibinfo{author}{A.~Holtzman}, \bibinfo{author}{L.~Zettlemoyer},
\newblock \bibinfo{title}{{QLoRA: Efficient finetuning of quantized LLMs}},
\newblock in: \bibinfo{booktitle}{NeurIPS}, \bibinfo{year}{2024}.
\bibitem[{Collarana et~al.(2024)Collarana, Busch, and
  Lange}]{collarana2024knowledge}
\bibinfo{author}{D.~Collarana}, \bibinfo{author}{M.~Busch},
  \bibinfo{author}{C.~Lange},
\newblock \bibinfo{title}{Knowledge graph treatments for hallucinating large
  language models},
\newblock \bibinfo{journal}{{ERCIM} News} \bibinfo{volume}{2024}
  (\bibinfo{year}{2024}).
\bibitem[{Pan et~al.(2024)Pan, Luo, Wang, Chen, Wang, and Wu}]{pan2024unifying}
\bibinfo{author}{S.~Pan}, \bibinfo{author}{L.~Luo}, \bibinfo{author}{Y.~Wang},
  \bibinfo{author}{C.~Chen}, \bibinfo{author}{J.~Wang},
  \bibinfo{author}{X.~Wu},
\newblock \bibinfo{title}{{Unifying Large Language Models and Knowledge Graphs:
  A Roadmap}},
\newblock \bibinfo{journal}{IEEE T-KDE}  (\bibinfo{year}{2024}).
\bibitem[{Shen et~al.(2023)Shen, Chang, Dong, Luo, and
  Meng}]{shen2023efficient}
\bibinfo{author}{H.~Shen}, \bibinfo{author}{H.~Chang},
  \bibinfo{author}{B.~Dong}, \bibinfo{author}{Y.~Luo},
  \bibinfo{author}{H.~Meng},
\newblock \bibinfo{title}{{Efficient LLM Inference on CPUs}},
\newblock in: \bibinfo{booktitle}{NeurIPS}, \bibinfo{year}{2023}.
\bibitem[{Bender et~al.(2021)Bender, Gebru, McMillan-Major, and
  Shmitchell}]{bender2021dangers}
\bibinfo{author}{E.~M. Bender}, \bibinfo{author}{T.~Gebru},
  \bibinfo{author}{A.~McMillan-Major}, \bibinfo{author}{S.~Shmitchell},
\newblock \bibinfo{title}{{On the dangers of stochastic parrots: Can language
  models be too big?}},
\newblock in: \bibinfo{booktitle}{ACM FAcct}, \bibinfo{year}{2021}.
\bibitem[{Curry et~al.(2023)Curry, Zaarour, Yang, Timilsina, Al-Qatf, and
  Haque}]{curry2023foundation}
\bibinfo{author}{E.~Curry}, \bibinfo{author}{T.~Zaarour},
  \bibinfo{author}{Y.~Yang}, \bibinfo{author}{M.~Timilsina},
  \bibinfo{author}{M.~Al-Qatf}, \bibinfo{author}{R.~Haque},
\newblock \bibinfo{title}{{Foundation Data Space Models: Bridging the
  Artificial Intelligence and Data Ecosystems (Vision Paper)}},
\newblock in: \bibinfo{booktitle}{2023 IEEE International Conference on Big
  Data (BigData)}, \bibinfo{organization}{IEEE}, \bibinfo{year}{2023}.

\end{thebibliography}

\end{document}